\documentclass[pmlr,twocolumn,10pt]{jmlr} % W&CP article

\makeatletter
\@ifundefined{mlhtrack}{%
  \newif\ifmlhneedspmlr \newif\ifmlhfindings \newif\ifmlhdemo
  \newif\ifmlhneedsstatements \newif\ifmlhanonymous
  \newcommand{\mlhtrack}[1]{\mlhneedspmlrtrue\mlhneedsstatementstrue\def\mlhtrackname{Proceedings Track}}%
}{}
\makeatother

\mlhtrack{proceedings}

\newif\iffinal
\finaltrue     % camera-ready version after acceptance

\iffinal
    \ifmlhneedspmlr
      \jmlrvolume{346}
      \jmlryear{2026}
    \fi
    \ifmlhfindings \jmlrproceedings{}{ML4H 2026 - Findings Track}\fi
    \ifmlhdemo     \jmlrproceedings{}{ML4H 2026 - Demo Track}\fi
    \jmlrworkshop{Machine Learning for Health (ML4H) 2026}
\else
    \jmlrproceedings{}{Submitted to ML4H 2026: \mlhtrackname}
    \jmlrworkshop{Machine Learning for Health (ML4H) 2026}
\fi

\usepackage{booktabs}
\usepackage{tikz}
\usetikzlibrary{arrows.meta,positioning,fit,backgrounds,calc}
\usepackage{pgfplots}\pgfplotsset{compat=1.16}
\usepackage{xcolor}
\usepackage{adjustbox}
\usepackage{microtype}

\definecolor{caccent}{HTML}{0F6E5C}
\definecolor{cclay}{HTML}{BD5B36}
\definecolor{cpaper}{HTML}{F3F5F3}
\definecolor{cline}{HTML}{C5CDC8}
\definecolor{cmuted}{HTML}{5C6B64}
\definecolor{ckhaki}{HTML}{7A6A3A}
\definecolor{hgreen}{HTML}{CDEBD9}
\definecolor{hyellow}{HTML}{F6E7A6}
\definecolor{hred}{HTML}{F3C9C4}
\newcommand{\oldv}[1]{\textcolor{cclay}{#1}}
\newcommand{\newv}[1]{\textcolor{caccent}{\textbf{#1}}}
\newcommand{\hg}[1]{\colorbox{hgreen}{#1}}
\newcommand{\hy}[1]{\colorbox{hyellow}{#1}}
\newcommand{\hr}[1]{\colorbox{hred}{#1}}
\newcommand{\arr}{\,$\rightarrow$\,\allowbreak}
\newcommand{\tp}[1]{\texttt{#1}}
\title[Surrogate Substitution Preserves PHI Detectability]{Surrogate Substitution Preserves PHI Detectability:\\ A Multi-Detector Equivalence Study}

\author{%
 \Name{Qiming Bao} \\
 \Name{Sherry J. H. Feng} \Email{sherry@custodianlabs.io}\\
 \Name{Kim Chester V. Eugenio} \\
 \Name{Meng Foon} \\
 \addr Custodian Labs
}

\begin{document}
\maketitle

\begin{abstract}
Structure-preserving de-identification replaces protected health information (PHI) with realistic same-type \emph{surrogates} (``Anna~S.'' becomes ``Maria~S.'', not \tp{[NAME]}). De-identified notes are what allow clinical text to be shared for model development, and many pipelines run a second PHI detector over them as a safety net. We ask a narrow, testable question: \emph{on the spans a de-identifier actually replaces, can downstream PHI detectors still find the surrogate as well as they found the original value?} We call this \emph{detectability preservation} and treat it as a proxy, not as general utility. We propose a paired protocol that (i)~scores the same spans before and after replacement with an identical panel of 11 detectors from three families, (ii)~measures the substitution effect only on replaced spans, keeping it separate from coverage, and (iii)~bounds the effect with document-level cluster-bootstrap equivalence testing and posterior probabilities instead of significance tests. Applied to a commercial transform on seven benchmarks in seven languages (1{,}750 documents), pooled recall on replaced spans moves from 75.9\% to 74.7\% ($\Delta=-1.24$ points, 90\% CI $[-1.65,-0.84]$), within a $\pm2$-point margin under overlap matching. The effect is not uniform: it exceeds $\pm2$ points for several detectors, for two identifier-dense non-English benchmarks, and under exact-boundary matching. Two-rater annotation of surrogates ($\kappa=0.79$), a same-slot counterfactual, and detector test--retest show that the residual has both generator-side and detector-side causes. On ICD-10 diagnosis classification of the same clinical documents, accuracy changes by at most 1 point under substitution, Faker surrogates, or redaction.
\end{abstract}
\begin{keywords}
De-identification, protected health information, surrogate generation, equivalence testing, clinical NLP
\end{keywords}

\ifmlhneedsstatements
\paragraph*{Data and Code Availability}
All seven benchmarks are public or synthetic: ASQ-PHI~\citep{weatherhead2026asqphi}, MEDDOCAN~\citep{marimon2019meddocan}, MultiCoNER~v2~\citep{fetahu2023multiconer}, and the English, Dutch, French and German subsets of PII-Masking-300k~\citep{ai4privacy2023}; ICD-10 labels for the MEDDOCAN documents come from the public CodiEsp corpus~\citep{mirandaescalada2020codiesp}. The 250-document evaluation subsets, their transformed versions, every detector's per-span predictions in both conditions, the two-rater annotations, and all analysis scripts are available at \url{https://github.com/Custodian-Labs/guardian-layer-phi-benchmark}.

\paragraph*{Institutional Review Board (IRB)}
This research does not require IRB approval. It uses only public or synthetic text and no real patient records, and the human annotation was performed by the authors on synthetic and public data.
\fi

\section{Introduction}
De-identified clinical notes are what make clinical text shareable for model training and evaluation under HIPAA and GDPR~\citep{hhs2012safeharbor}. De-identification takes two broad forms. \textbf{Redaction} deletes or tags PHI (\oldv{\tp{John Smith}}\arr\tp{[NAME]}); redacted corpora such as the MIMIC notes are widely and successfully used, and whether redaction costs downstream performance depends on the task. \textbf{Structure-preserving} de-identification substitutes each PHI value with a plausible same-type surrogate (\oldv{\tp{John Smith}}\arr\newv{\tp{Maria Lopez}}), keeping the text fluent~\citep{carrell2013hiding}. Substitution is established practice: the i2b2/UTHealth 2014 corpus~\citep{stubbs2015i2b2} and MEDDOCAN~\citep{marimon2019meddocan} were released with realistic surrogate PHI, pseudonymised clinical text has been used to pre-train clinical language models~\citep{vakili2022downstream}, and open toolkits such as Presidio~\citep{presidio} support replacement as well as redaction. We know of no survey of how often it is deployed.

Within this setting, \emph{detectability parity} between surrogates and real PHI matters in two concrete ways. ``Hiding in plain sight''~\citep{carrell2013hiding} protects residual, unmasked PHI only if surrogates cannot be told apart from real values; a surrogate that detectors treat differently weakens that protection. And layered de-identification runs a second-pass detector after substitution, which only works if the second pass still sees surrogates as PHI. We therefore ask: \emph{given that a span was replaced, does a downstream detector find the surrogate as often as it found the original value?}

Answering this needs three things that existing evaluations often combine. Coverage (how much PHI the transform replaces) can be compared on whole-document metrics, but the effect of substitution on the replaced spans needs a separate, paired measurement. With tens of thousands of paired outcomes, a significance test rejects for negligible effects~\citep{dror2018hitchhiker,card2020power}, so a bounded-effect statement needs equivalence testing~\citep{schuirmann1987tost,lakens2017tost}. And an aggregate change does not say whether it comes from the surrogates, the detectors, or the evaluation itself.

\textbf{Contributions.} We do not propose new statistical techniques. Our contribution is a protocol that separates three quantities: (i)~\emph{coverage}; (ii)~the \emph{per-span substitution effect}, measured paired on the same replaced spans across a heterogeneous detector panel and bounded with clustering-aware equivalence testing; and (iii)~its \emph{mechanism}, using human-validated surrogate labels, a same-slot counterfactual, and detector test--retest. We apply it to one commercial transform as a case study, with open baselines (redaction, Faker surrogates) for reproducibility, and add one clinical downstream task.

\section{Related Work}
\textbf{Surrogates and detection.} Clinical de-identification is usually framed as sequence labelling~\citep{uzuner2007evaluating,stubbs2015i2b2,marimon2019meddocan}, with systems from rule-based pipelines~\citep{presidio} to transformer taggers~\citep{johnson2020deid} and zero-shot LLMs~\citep{liu2023deidgpt}. The work closest to ours is \citet{yeniterzi2010resynthesis}, who retrained one CRF de-identifier on original and resynthesised records: training on original data and testing on resynthesised data lowered F from 0.960 to 0.862, mostly through recall (0.978 to 0.816). \citet{deleger2014preparing} and \citet{eder2019deidentification} report related train/test comparisons on surrogate corpora. These studies measure train/test distribution shift with one retrained tagger. Our setting is the inference-time counterpart: fixed, off-the-shelf detectors from three families, no retraining, the same spans compared before and after replacement, and seven languages. Surrogate quality also matters for privacy: poor surrogates can help an attacker separate real from fake identifiers~\citep{carrell2020parrot}.

\textbf{LLMs as detectors.} Nine of our eleven detectors are LLMs prompted zero-shot. Their behaviour depends on the prompt and on how reasoning is organised: structure-aware planning with a world model~\citep{xiong2025deliberate}, structured multi-level modelling~\citep{xiong2026enhancing}, multi-path plan aggregation for long chain-of-thought~\citep{xiong2025enhancing}, adaptive information control in search-augmented reasoning~\citep{xiong2026adaptive}, and diversity-driven self-improvement of frozen models~\citep{xiong2026dive} all change what an LLM produces for the same input. We fix one prompt and decoding setting per model, so our estimates describe the substitution effect for these configurations; the paired design applies unchanged to stronger reasoning strategies.

\textbf{Downstream utility.} \citet{vakili2022downstream} pre-train BERT on automatically pseudonymised Swedish clinical text and find no drop on six downstream tasks; \citet{berg2020impact} report a low impact of pseudonymisation on clinical NER. A null result from one downstream model is consistent both with surrogates preserving the signal and with a task that is insensitive to the small fraction of tokens that are PHI, and it does not show which spans changed. Our paired, per-span design addresses this, and \S\ref{sec:downstream} adds a diagnosis-classification task.

\textbf{Equivalence testing.} The two one-sided tests (TOST) procedure~\citep{schuirmann1987tost,berger1996bioequivalence} is the standard way to establish bioequivalence~\citep{fda2001bioequivalence} and is covered by the CONSORT extension for non-inferiority and equivalence trials~\citep{piaggio2012consort,walker2011equivalence}; \citet{lakens2017tost} give a tutorial. It is less common in NLP evaluation. We pair it with a Bayesian bootstrap~\citep{rubin1981bayesian}.

\section{Protocol}\label{sec:protocol}
\begin{figure*}[t]
\centering
\begin{adjustbox}{max width=\textwidth}
\begin{tikzpicture}[font=\scriptsize,>={Stealth[length=2.2mm]},
  card/.style={draw=cline,fill=white,rounded corners=2pt,inner sep=4pt,align=left,text width=3.55cm},
  op/.style={draw=caccent,fill=caccent!12,rounded corners=2pt,inner sep=3pt,align=center,minimum height=6mm,font=\scriptsize\bfseries},
  det/.style={draw=caccent,fill=caccent!12,rounded corners=3pt,inner sep=4pt,align=center,text width=2.05cm,minimum height=17mm},
  meas/.style={draw=cline,fill=white,rounded corners=2pt,inner sep=5pt,align=center,text width=2.6cm},
  verdict/.style={draw=caccent,line width=0.8pt,fill=caccent!14,rounded corners=3pt,inner sep=5pt,align=center,text width=2.9cm},
  flow/.style={->,thick,caccent!85},
  panel/.style={rounded corners=3pt,fill=black!3,draw=cline!70,inner sep=8pt},
  badge/.style={circle,fill=caccent,text=white,font=\scriptsize\bfseries,inner sep=0.6pt,minimum size=3.8mm}
]
\node[card] (co) at (0,1.25) {\dots a 34-year-old female like \oldv{Anna S.}, seen at Methodist Hospital on \oldv{April 12, 2023}?};
\node[op]   (T)  at (0,0)    {Transform $T$};
\node[card] (ct) at (0,-1.25){\dots a 35-year-old female like \newv{Maria S.}, seen at Methodist Hospital on \newv{March 13, 2021}?};
\draw[flow] (co.south) -- (T.north);
\draw[flow] (T.south) -- (ct.north);
\node[font=\scriptsize\itshape,text=cmuted,right=1pt of T,text width=1.5cm] {same-type surrogate};
\node[det] (det) at (5.15,0) {Identical\\ detector\\ panel $D$\\ ($\times\,11$)};
\draw[flow] (co.east) to[out=0,in=145] (det.north west);
\draw[flow] (ct.east) to[out=0,in=215] (det.south west);
\node[meas] (m) at (8.55,0) {recall on replaced spans\\[2pt]\textbf{75.9\%}\,$\rightarrow$\,\textbf{74.7\%}\\{\scriptsize($\Delta{=}{-}1.24$ pts; 5{,}273 spans)}};
\draw[flow] (det.east) -- (m.west);
\node[verdict] (v) at (12.0,0) {cluster bootstrap\\ 90\% CI $[-1.65,-0.84]$\\[2pt]$P(|\Delta|{<}2)>0.99$\\[3pt]{\color{caccent}\textbf{within $\pm2$ (pooled, overlap)}}};
\draw[flow] (m.east) -- (v.west);
\begin{scope}[on background layer]
  \node[panel,fit=(co)(ct)(T)] (p1) {};
  \node[panel,fit=(det)] (p2) {};
  \node[panel,fit=(m)] (p3) {};
  \node[panel,fit=(v)] (p4) {};
\end{scope}
\foreach \p/\n/\ttl in {p1/1/Substitute,p2/2/Detect (paired),p3/3/Measure,p4/4/Bound the effect}{
  \node[badge] (b\p) at ([xshift=5pt,yshift=6pt]\p.north west) {\n};
  \node[anchor=west,font=\scriptsize\bfseries,text=black] at (b\p.east) {\,\ttl};
}
\end{tikzpicture}
\end{adjustbox}
\caption{The paired protocol with a worked example. (1)~The transform replaces each \emph{detected} PHI value with a same-type surrogate. (2)~The identical 11-detector panel scores the original $d$ and the transformed $d'$. (3)~Recall is measured only on the replaced spans, separately from coverage. (4)~The paired change is bounded with a document-level cluster bootstrap and posterior probabilities. The ages in the example are not gold PHI on this benchmark, so they are not scored.}
\label{fig:protocol}
\end{figure*}

\textbf{Definitions.} A transform $T$ maps document $d$ to $d'$, replacing each PHI value $v$ it detects with a surrogate $s$. The following are our operational definitions. \emph{Coverage} is the share of gold PHI spans whose text $T$ changed. \emph{Detectability preservation} concerns the replaced spans only: for each replaced gold span and each detector, we record whether the detector finds $v$ in $d$ and $s$ in $d'$, and we report
\[\Delta=\mathrm{recall}(d')-\mathrm{recall}(d)\]
in percentage points, so negative values are losses. We use this one change metric throughout.

\textbf{Re-projection.} Gold spans are aligned onto $d'$ by character-level alignment (difflib). Where the original span ends on a word boundary, we extend the aligned span on $d'$ outward to the nearest word boundary. This corrects alignments that cut through a surrogate (\tp{1[0/03/2032]}) and recovers replacements that the alignment had truncated to the unchanged prefix (\tp{Colombia}\arr\tp{Colombian}). The transform's own replacement offsets were not available, so this is a heuristic correction. The uncorrected alignment gives nearly the same headline ($\Delta=-1.22$; Table~\ref{tab:criteria}). Gold spans whose stored text does not match the document (59, from template placeholders in PII-Masking-300k) are excluded.

\textbf{Matching.} Detectors emit incompatible label sets (Presidio entity types, i2b2 types, free-text LLM labels), so the primary criterion is label-agnostic \emph{overlap}: a span counts as found when any prediction shares at least one character with it. This asks whether the detector found the PHI, which is the privacy-relevant question. Table~\ref{tab:matching} shows the criteria on an example, and Table~\ref{tab:criteria} reports results under IoU $\geq0.5$, exact boundaries, and exact boundaries with a coarse shared type map.

\begin{table}[t]\centering\small
\caption{Matching criteria for the gold span \tp{John L.} (NAME) in ``\dots seen by Dr.~John~L. at Mt.~Sinai\dots''.}\label{tab:matching}
\begin{adjustbox}{max width=\columnwidth}
\begin{tabular}{lccc}
\toprule
Prediction & Exact+type & Exact & Overlap\\
\midrule
\tp{John L.} as NAME & \checkmark & \checkmark & \checkmark\\
\tp{John L.} as PERSON & \checkmark & \checkmark & \checkmark\\
\tp{John L.} as DATE & -- & \checkmark & \checkmark\\
\tp{Dr. John} & -- & -- & \checkmark\\
\tp{Mt. Sinai} only & -- & -- & --\\
\bottomrule
\end{tabular}
\end{adjustbox}
\end{table}

\textbf{Inference.} Each benchmark contributes many span$\times$detector outcomes per document, which are correlated. We therefore resample documents within each benchmark (stratified cluster bootstrap, 5{,}000 draws) and compute the pooled ratio estimator of $\Delta$. A change is equivalent within margin $m$ (TOST, $\alpha=0.05$) iff its 90\% CI lies inside $(-m,m)$. We report the smallest such margin $m^\ast$ and, from a Bayesian bootstrap over documents~\citep{rubin1981bayesian}, the posterior probability $P(|\Delta|<m)$ for $m\in\{1,2,3\}$. The margin $m=2$ is a convention (about 2.6\% of a 76\% baseline). It is not derived from a downstream requirement, so we report $m^\ast$ and the margin curve (Appendix~\ref{app:margins}) and let readers apply their own tolerance.

\textbf{Result hierarchy.} \emph{Primary}: the pooled paired change on replaced spans. \emph{Secondary}: per-benchmark and per-detector changes; coverage and leakage; robustness to matching. \emph{Exploratory}: the redaction and Faker contrasts, the downstream task, and the mechanism analyses in \S\ref{sec:mechanism}.

\section{Experimental Design}\label{sec:design}
\textbf{Detector panel.} Eleven detectors in three families (details in Appendix~\ref{app:detectors}): a rule-based pipeline, Microsoft Presidio~\citep{presidio}; an encoder tagger, OBI \tp{deid\_roberta\_i2b2}~\citep{johnson2020deid}; and nine LLMs prompted zero-shot with one fixed prompt each: Gemma~4~31B and E4B~\citep{gemma2report}, Qwen~3.5 4B, 9B and 35B-A3B~\citep{qwen25report}, Llama~3.1~8B and 3.3~70B~\citep{llama3herd}, DeepSeek~V2-Lite~\citep{deepseekv2}, all run locally, and GPT-4o-mini~\citep{openai2024gpt4o} through the OpenAI API at temperature 0. We recover LLM character spans by string search over the source text. Gemma~4~31B lacks transformed outputs for 149 ASQ-PHI documents and Llama~3.3~70B for parts of MEDDOCAN and PII-nl; their per-detector rows use the documents they cover. We also ran GPT-5~\citep{openai2025gpt5} on a stratified subset (Appendix~\ref{app:gpt5}).

\textbf{Benchmarks.} Seven subsets of 250 documents each (1{,}750 in total): ASQ-PHI (synthetic English clinical queries)~\citep{weatherhead2026asqphi}, MEDDOCAN (Spanish clinical case reports with synthetic PHI)~\citep{marimon2019meddocan}, MultiCoNER~v2~\citep{fetahu2023multiconer}, and PII-Masking-300k in English, Dutch, French and German~\citep{ai4privacy2023}. They cover clinical and general PII text and both free-form and JSON formats.

\textbf{Transform.} The case study is the Guardian Layer de-identification service of Custodian Labs~\citep{custodian_labs} in \tp{transform} mode (top-1 surrogate, all entity types, \tp{domain=General}). The service is a black box to this evaluation; it does not use any model in our detector panel.

\textbf{Contrasts and additional analyses.} \emph{C1, redaction}: each replaced value is overwritten with \tp{*} characters and re-detected; we also run Qwen-9B with one added prompt sentence stating that runs of \tp{*} mark removed PHI. \emph{C2, open surrogates}: each replaced value is substituted with a seeded, same-type, locale-matched Faker~\citep{faker} value. C1 and C2 use four detectors spanning the three families on the first 120 documents per benchmark, because of the compute cost of re-scoring every corpus. \emph{Downstream}: \S\ref{sec:downstream}. \emph{Mechanism}: \S\ref{sec:mechanism}.

\section{Results}\label{sec:results}
\textbf{Primary result.} Over 5{,}273 replaced gold spans in 1{,}142 documents (57{,}035 span$\times$detector outcomes), pooled recall moves from 75.9\% to 74.7\% (Table~\ref{tab:bench}): $\Delta=-1.24$ points, 90\% cluster-bootstrap CI $[-1.65,-0.84]$, smallest passing margin $m^\ast=1.65$. The posterior probability that $|\Delta|<2$ exceeds 0.99, but $P(|\Delta|<1)=0.15$, so we do not claim that substitution has no effect. For a team that substitutes PHI and then runs a detector as a safety net, the share of replaced spans the detector misses rises from 24.1\% to 25.3\%. This concerns detectability of replaced spans only; overall PHI exposure also depends on coverage.

\begin{table*}[t]\centering\small
\caption{Per-benchmark results on replaced spans (overlap matching). Cov.\ = coverage, the share of gold PHI spans the transform replaced (\%). Outc.\ = span$\times$detector outcomes. $\Delta$ = transformed $-$ original recall (points). $m^\ast$ = smallest margin at which TOST passes. $P$ = posterior probability from a Bayesian bootstrap over documents.}\label{tab:bench}
\begin{tabular}{lrrrrrr}
\toprule
Benchmark & Cov. & Outc. & Recall O$\rightarrow$T & $\Delta$ [90\% CI] & $m^\ast$ & $P(|\Delta|{<}2)$\\
\midrule
ASQ-PHI & 77.8 & 5{,}457 & 91.9$\rightarrow$91.1 & $-$0.81 [$-$1.36, $-$0.24] & 1.36 & $>$0.99\\
MEDDOCAN & 52.2 & 32{,}172 & 73.3$\rightarrow$71.4 & $-$1.88 [$-$2.43, $-$1.30] & 2.43 & 0.64\\
MultiCoNER & 5.7 & 220 & 71.8$\rightarrow$71.4 & $-$0.45 [$-$5.19, $+$4.31] & 5.19 & 0.56\\
PII-en & 32.0 & 5{,}280 & 75.3$\rightarrow$74.3 & $-$0.97 [$-$1.97, $+$0.02] & 1.97 & 0.96\\
PII-nl & 22.0 & 3{,}808 & 77.2$\rightarrow$75.3 & $-$1.86 [$-$3.08, $-$0.60] & 3.08 & 0.58\\
PII-fr & 30.0 & 5{,}720 & 75.6$\rightarrow$76.4 & $+$0.86 [$-$0.73, $+$2.80] & 2.80 & 0.87\\
PII-de & 26.0 & 4{,}378 & 76.0$\rightarrow$76.3 & $+$0.30 [$-$1.04, $+$1.61] & 1.61 & 0.98\\
\midrule
\textbf{Pooled} & & \textbf{57{,}035} & \textbf{75.9$\rightarrow$74.7} & \textbf{$-$1.24} [$-$1.65, $-$0.84] & \textbf{1.65} & $>$0.99\\
\bottomrule
\end{tabular}

\end{table*}

\textbf{The effect is not uniform} (Figure~\ref{fig:forest}). Clean English text stays well within $\pm2$ points (ASQ-PHI $m^\ast=1.36$, PII-de 1.61, PII-en 1.97). The identifier-dense non-English benchmarks do not: MEDDOCAN needs $\pm2.43$ and PII-nl $\pm3.08$, each with only about a 60\% posterior probability of lying within $\pm2$, and PII-fr needs $\pm2.80$. MultiCoNER has too few replaced spans (20) for a useful bound. By detector (Table~\ref{tab:det}), only Gemma~4~31B ($m^\ast=0.70$) and Qwen~3.5~9B (1.01) are equivalent within $\pm2$ on their own; OBI loses 2.69 points, DeepSeek~V2-Lite 4.02, and the weak Llama~3.1~8B gains 2.84. The pooled result therefore does not imply equivalence for every benchmark or detector.

\begin{figure}[t]
\centering
\begin{adjustbox}{max width=\columnwidth}
\begin{tikzpicture}
\begin{axis}[
  width=\linewidth, height=5.0cm,
  xmin=-5.6, xmax=5.0, xlabel={$\Delta$: transformed $-$ original recall (pts)},
  xlabel style={font=\scriptsize}, tick label style={font=\scriptsize},
  ytick={1,2,3,4,5,6,7,8},
  yticklabels={Pooled,PII-de,PII-fr,PII-en,ASQ-PHI,MultiCoNER,PII-nl,MEDDOCAN},
  ymin=0.4, ymax=8.9, axis y line=left, axis x line=bottom, clip=false,
]
\fill[caccent!12] (axis cs:-2,0.4) rectangle (axis cs:2,8.6);
\draw[caccent!65,dashed] (axis cs:-2,0.4)--(axis cs:-2,8.6);
\draw[caccent!65,dashed] (axis cs:2,0.4)--(axis cs:2,8.6);
\draw[black!45] (axis cs:0,0.4)--(axis cs:0,8.6);
\addplot[only marks,mark=*,mark size=1.5pt,color=cclay,
  error bars/.cd,x dir=both,x explicit] coordinates {
  (-1.24,1) += (0.40,0) -= (0.41,0)
  (0.30,2)  += (1.31,0) -= (1.34,0)
  (0.86,3)  += (1.94,0) -= (1.59,0)
  (-0.97,4) += (0.99,0) -= (1.00,0)
  (-0.81,5) += (0.57,0) -= (0.55,0)
  (-0.45,6) += (4.76,0) -= (4.74,0)
  (-1.86,7) += (1.26,0) -= (1.22,0)
  (-1.88,8) += (0.58,0) -= (0.55,0)};
\node[font=\scriptsize,text=caccent,anchor=south] at (axis cs:0,8.6) {margin $\pm2$};
\end{axis}
\end{tikzpicture}
\end{adjustbox}
\caption{$\Delta$ per benchmark with 90\% cluster-bootstrap CIs (TOST at $\alpha=0.05$ passes iff the interval lies inside the band).}
\label{fig:forest}
\end{figure}

\begin{table*}[t]\centering\small
\caption{Per detector. Whole document (all gold PHI, mean over benchmarks): span F1 with exact boundaries, and leakage = share of gold PHI spans not overlapped by any prediction (\%). Replaced spans: recall, $\Delta$ (points), 90\% cluster-bootstrap CI and $m^\ast$. $^\dagger$Incomplete run (see \S\ref{sec:design}). The low whole-document F1 of OBI reflects boundary fragmentation, not missed PHI (its leakage is 30\%).}\label{tab:det}
\begin{tabular}{lrrrrrr}
\toprule
 & \multicolumn{2}{c}{Whole document} & \multicolumn{4}{c}{Masked spans}\\
\cmidrule(lr){2-3}\cmidrule(lr){4-7}
Detector & F1 O$\rightarrow$T & Leak O$\rightarrow$T & Recall O$\rightarrow$T & $\Delta$ & 90\% CI & $m^\ast$\\
\midrule
Gemma 4 31B$^\dagger$ & .744$\rightarrow$.735 & 17.2$\rightarrow$17.0 & 91.9$\rightarrow$91.9 & $+$0.08 & [$-$0.51, $+$0.70] & 0.70\\
Llama 3.3 70B$^\dagger$ & .725$\rightarrow$.719 & 13.9$\rightarrow$13.6 & 90.6$\rightarrow$89.2 & $-$1.40 & [$-$2.15, $-$0.64] & 2.15\\
Qwen 3.5 35B-A3B & .715$\rightarrow$.693 & 14.5$\rightarrow$14.9 & 86.6$\rightarrow$85.0 & $-$1.56 & [$-$2.56, $-$0.57] & 2.56\\
GPT-4o-mini & .706$\rightarrow$.696 & 19.6$\rightarrow$19.6 & 89.3$\rightarrow$87.4 & $-$1.92 & [$-$2.76, $-$1.10] & 2.76\\
Gemma 4 E4B & .737$\rightarrow$.725 & 18.7$\rightarrow$18.9 & 87.3$\rightarrow$85.4 & $-$1.90 & [$-$2.73, $-$1.07] & 2.73\\
Qwen 3.5 9B & .655$\rightarrow$.647 & 30.3$\rightarrow$29.9 & 79.7$\rightarrow$79.7 & $+$0.06 & [$-$0.88, $+$1.01] & 1.01\\
Qwen 3.5 4B & .568$\rightarrow$.554 & 38.6$\rightarrow$38.8 & 65.9$\rightarrow$64.8 & $-$1.10 & [$-$2.72, $+$0.48] & 2.72\\
Presidio & .415$\rightarrow$.414 & 38.0$\rightarrow$38.5 & 74.9$\rightarrow$72.9 & $-$2.01 & [$-$2.73, $-$1.30] & 2.73\\
DeepSeek V2-Lite & .411$\rightarrow$.401 & 56.4$\rightarrow$57.7 & 51.4$\rightarrow$47.4 & $-$4.02 & [$-$5.69, $-$2.37] & 5.69\\
Llama 3.1 8B & .392$\rightarrow$.392 & 53.1$\rightarrow$53.1 & 47.4$\rightarrow$50.2 & $+$2.84 & [$+$0.51, $+$5.27] & 5.27\\
OBI RoBERTa & .041$\rightarrow$.042 & 29.9$\rightarrow$31.0 & 73.3$\rightarrow$70.6 & $-$2.69 & [$-$3.27, $-$2.10] & 3.27\\
\bottomrule
\end{tabular}

\end{table*}

\textbf{Robustness to matching.} Under stricter criteria the loss is about 2 points: $-1.62$ with IoU $\geq0.5$, $-1.94$ with exact boundaries and $-2.13$ with exact boundaries and type (Table~\ref{tab:criteria}), with $m^\ast$ between 2.03 and 2.56. Equivalence within $\pm2$ is therefore established only under overlap matching. Most of the extra boundary loss in the uncorrected alignment came from the alignment itself: before correction, 7.1\% of re-projected replaced spans cut through a word on $d'$, against 0.4\% of the same spans on $d$.

\begin{table}[t]\centering\small
\caption{Pooled $\Delta$ under each matching criterion (corrected re-projection), and under overlap with the uncorrected difflib alignment.}\label{tab:criteria}
\begin{adjustbox}{max width=\columnwidth}
\begin{tabular}{lrrr}
\toprule
Criterion & O$\rightarrow$T & $\Delta$ [90\% CI] & $m^\ast$\\
\midrule
Overlap (primary) & 75.9$\rightarrow$74.7 & $-$1.24 [$-$1.65, $-$0.84] & 1.65\\
IoU $\geq$ 0.5 & 68.4$\rightarrow$66.8 & $-$1.62 [$-$2.03, $-$1.20] & 2.03\\
Exact & 59.8$\rightarrow$57.8 & $-$1.94 [$-$2.40, $-$1.50] & 2.40\\
Exact + type & 56.8$\rightarrow$54.7 & $-$2.13 [$-$2.56, $-$1.71] & 2.56\\
\midrule
Overlap, uncorrected & 76.1$\rightarrow$74.9 & $-$1.22 [$-$1.62, $-$0.83] & 1.62\\
\bottomrule
\end{tabular}

\end{adjustbox}
\end{table}

\textbf{Coverage and leakage.} Coverage is a privacy outcome of its own and is far from complete: the transform replaced 77.8\% of gold PHI spans on ASQ-PHI and 52.2\% on MEDDOCAN, and 6--32\% on the general-domain corpora, whose annotated entities (encyclopaedic names, generic places, usernames) fall partly outside its scope (Table~\ref{tab:bench}). At this coverage the transform is not sufficient as a standalone de-identifier. Whole-document leakage changes by $-0.4$ to $+1.3$ points per detector and whole-document F1 by $-0.022$ to $+0.001$, and detector ranking is preserved (Table~\ref{tab:det}).

\textbf{Span selection.} Because the transform chooses which spans to replace, the replaced population is not representative of all PHI: on the original documents, detectors find 75.9\% of replaced gold spans but only 59.3\% of the gold spans the transform left in place, with the largest gap for dates (83.5\% vs.\ 43.0\%; Appendix~\ref{app:selection}). The paired $\Delta$ is internally valid, but absolute recall levels describe the replaced population only. Reweighting the per-(benchmark, type) changes to the type composition of all gold PHI gives $\Delta=-0.66$, so selection on type does not hide a larger loss; it cannot correct for selection within a type.

\textbf{Redaction and open surrogates (C1, C2).} With \tp{*****} in place of the value, Presidio and OBI find 0--5\% of replaced spans, while LLMs partly recover them (Qwen-9B 6--58\%, Gemma-31B 13--80\%; Table~\ref{tab:c1c2}). This floor applies only to detectors not adapted to the mask token: adding one prompt sentence about \tp{*} runs raises Qwen-9B to 55--77\%, and a regular expression would reach nearly 100\%. Redaction is therefore a reasonable choice whenever downstream use does not need the entity signal. Faker surrogates keep recall high in every family (means 77--94\%), often above the original values, because canonical values are easier to detect than real PHI; Faker is therefore not a neutral control. On the same spans, the original/commercial/Faker recall is 75.9/73.7/78.9\% for Presidio, 72.4/70.1/74.1\% for OBI and 78.4/78.5/82.4\% for Qwen-9B.

\begin{table*}[t]\centering\small
\caption{C1 and C2: recall (\%) on replaced spans after redaction or Faker substitution, first 120 documents per benchmark. $N$ = replaced spans. $^\ddagger$Qwen-9B with one prompt sentence stating that runs of \tp{*} mark removed PHI. MultiCoNER (8 spans) is illustrative only.}\label{tab:c1c2}
\begin{adjustbox}{max width=\textwidth}
\begin{tabular}{lrrrrrrrrrr}
\toprule
 & & \multicolumn{5}{c}{C1: redaction (\tp{*****})} & \multicolumn{4}{c}{C2: Faker surrogates}\\
\cmidrule(lr){3-7}\cmidrule(lr){8-11}
Benchmark & $N$ & Presidio & OBI & Qwen-9B & Gemma-31B & Qwen-9B$^{\ddagger}$ & Presidio & OBI & Qwen-9B & Gemma-31B\\
\midrule
ASQ-PHI & 267 & 0.0 & 0.7 & 58.1 & 79.8 & 76.0 & 96.3 & 100.0 & 100.0 & 100.0\\
MEDDOCAN & 1{,}419 & 0.0 & 0.1 & 22.9 & 52.6 & 54.9 & 78.4 & 64.0 & 79.6 & 96.5\\
MultiCoNER & 8 & 0.0 & 0.0 & 25.0 & 12.5 & 87.5 & 100.0 & 100.0 & 87.5 & 100.0\\
PII-en & 210 & 1.0 & 1.0 & 10.0 & 37.6 & 71.4 & 77.1 & 89.5 & 76.7 & 90.0\\
PII-nl & 195 & 5.1 & 1.0 & 20.5 & 39.5 & 74.4 & 76.4 & 71.8 & 84.1 & 96.9\\
PII-fr & 220 & 3.6 & 0.5 & 5.9 & 38.2 & 76.8 & 65.9 & 79.5 & 81.4 & 83.2\\
PII-de & 204 & 0.0 & 1.0 & 11.8 & 41.7 & 73.0 & 67.2 & 84.8 & 77.9 & 95.1\\
\midrule
Mean (excl.\ MultiCoNER) & & 1.6 & 0.7 & 21.5 & 48.2 & 71.1 & 76.9 & 81.6 & 83.3 & 93.6\\
\bottomrule
\end{tabular}

\end{adjustbox}
\end{table*}

\subsection{Downstream: ICD-10 Diagnosis Classification}\label{sec:downstream}
Detectability is a narrow proxy. As one clinical downstream check, we use the ICD-10 diagnosis codes that the CodiEsp corpus~\citep{mirandaescalada2020codiesp} provides for all 250 MEDDOCAN case reports; these labels do not depend on the PHI. Three open LLMs (Qwen~3.5 4B and 9B, Gemma~4~E4B) answer, for each of eight ICD-10 chapters with 20--40\% prevalence, whether the patient has a diagnosis in that chapter. Answers are scored by option likelihood, so no output parsing is needed, and each model is compared with itself on the original text. Diagnosis classification is unaffected by every de-identification method: macro accuracy changes by $-0.4$ to $+0.5$ points and AUROC by $-0.3$ to $+0.4$, with all 95\% document-bootstrap intervals within $\pm1$ point (Table~\ref{tab:downstream}). The methods differ in the clinical consistency of the replaced values: against the MEDDOCAN annotation of patient age, age-group accuracy changes by at most 2.4 points under the commercial transform, by 5--7 points under redaction, and by 14--23 points under Faker, which draws ages at random. This is one task family with small local models; it does not establish general clinical utility.

\begin{table*}[t]\centering\small
\caption{ICD-10 chapter classification on MEDDOCAN (CodiEsp labels, 8 chapters), change in macro accuracy and AUROC against the same model on the original text (points; 95\% document-bootstrap CI). Sex and age group: change in accuracy against the MEDDOCAN annotation.}\label{tab:downstream}
\begin{tabular}{llrrrr}
\toprule
Model & Text & ICD-10 acc.\ $\Delta$ [95\% CI] & AUROC $\Delta$ & Sex $\Delta$ & Age group $\Delta$\\
\midrule
Qwen 3.5 4B & Guardian Layer & $+$0.3 [$-$0.1, $+$0.9] & $-$0.2 & $+$0.8 & $-$2.4\\
 & Faker & $+$0.2 [$-$0.3, $+$0.6] & $+$0.1 & $-$1.2 & $-$18.1\\
 & Redacted & $-$0.1 [$-$0.5, $+$0.3] & $-$0.1 & $-$0.4 & $-$7.3\\
\addlinespace
Qwen 3.5 9B & Guardian Layer & $+$0.5 [$-$0.0, $+$1.0] & $-$0.3 & $+$0.4 & $-$1.2\\
 & Faker & $-$0.1 [$-$0.4, $+$0.3] & $-$0.3 & $-$0.8 & $-$14.1\\
 & Redacted & $+$0.0 [$-$0.4, $+$0.4] & $-$0.1 & $-$6.5 & $-$5.2\\
\addlinespace
Gemma 4 E4B & Guardian Layer & $-$0.2 [$-$0.7, $+$0.3] & $+$0.4 & $+$0.0 & $+$2.0\\
 & Faker & $-$0.4 [$-$0.9, $+$0.2] & $-$0.1 & $+$0.4 & $-$23.0\\
 & Redacted & $-$0.3 [$-$0.8, $+$0.2] & $+$0.2 & $-$0.4 & $-$6.9\\
\bottomrule
\end{tabular}

\end{table*}

\section{What Drives the Residual?}\label{sec:mechanism}
\textbf{Human-validated surrogate quality.} Two authors independently labelled 197 replaced surrogates (a stratified sample across the seven benchmarks), blind to detector outcomes and to an earlier AI-generated labelling, with eight categories (Appendix~\ref{app:annotation}). Agreement was substantial: Cohen's $\kappa=0.79$~\citep{cohen1960kappa} (88.8\% raw). Each rater agreed with the AI labels only at $\kappa=0.47$ and $0.49$, so we use the human labels only. By consensus, 65\% of surrogates are well-formed, 23\% are defective (x-masked 18, context-inconsistent 12, malformed or truncated 10, type-inconsistent 4, locale-inconsistent 1) and 11\% are disagreements. Among detections that succeeded on the original, defective surrogates are lost in 8.6\% of cases and well-formed ones in 5.1\% (95\% CI 3.7--6.6; Table~\ref{tab:human}). About half of all lost outcomes involve well-formed surrogates. Generator defects raise the loss rate, but they do not account for most of the residual. Producing well-formed text that stays consistent with its context is itself a hard generation problem~\citep{bao2024amr}, so some defect rate is expected from any generator.

\begin{table}[t]\centering\small
\caption{Consensus surrogate labels (197 spans) linked to all detectors' outcomes. Lost = found on the original, missed after transformation.}\label{tab:human}
\begin{adjustbox}{max width=\columnwidth}
\begin{tabular}{lrrr}
\toprule
Consensus label & Spans & Found on O & Lost after T\\
\midrule
Well-formed & 128 & 1058 & 54 (5.1\%)\\
x-masked & 18 & 156 & 13 (8.3\%)\\
Malformed / truncated & 10 & 69 & 5 (7.2\%)\\
Context-inconsistent & 12 & 75 & 7 (9.3\%)\\
Type-inconsistent & 4 & 27 & 3 (11.1\%)\\
Locale-inconsistent & 1 & 10 & 1 (10.0\%)\\
Rater disagreement & 22 & 169 & 18 (10.7\%)\\
\bottomrule
\end{tabular}

\end{adjustbox}
\end{table}

\textbf{Same-slot counterfactual.} For every outcome lost under the commercial surrogate, we place a Faker value in exactly the same slot and re-run the same detector (Presidio, OBI and Qwen-9B; 120 documents per benchmark). Of 335 lost outcomes, 71\% (95\% CI 65--77) are found with the Faker value (Table~\ref{tab:cf}); the rest are missed under both surrogates, pointing to the detector or the context. The reverse also holds: 78\% of outcomes lost under Faker are kept under the commercial surrogate. Loss is therefore mostly specific to the particular value drawn rather than to either generator, with a larger detector or context share for the LLM detector (60\% recovered).

\begin{table}[t]\centering\small
\caption{Same-slot counterfactual. Lost (T): outcomes found on the original and missed under the commercial surrogate; Found (Faker): of these, found when a Faker value fills the same slot; Kept by T: outcomes lost under Faker but found under the commercial surrogate.}\label{tab:cf}
\begin{adjustbox}{max width=\columnwidth}
\begin{tabular}{lrrr}
\toprule
Detector & Lost (T) & Found (Faker) & Kept by T\\
\midrule
Presidio & 124 & 99 (80\%) & 150/175 (86\%)\\
OBI RoBERTa & 99 & 72 (73\%) & 130/157 (83\%)\\
Qwen 3.5 9B & 112 & 67 (60\%) & 71/116 (61\%)\\
\textbf{All} & 335 & 238 (71\%) & 351/448 (78\%)\\
\bottomrule
\end{tabular}

\end{adjustbox}
\end{table}

\textbf{Detector test--retest.} OBI gives identical outputs when re-run on all seven benchmarks. GPT-4o-mini at temperature 0, re-run three times on 199 documents (751 replaced spans), flips 4.2\% (original) and 5.0\% (transformed) of span outcomes between runs, against 7.5--8.8\% between conditions; the estimated $\Delta$ is negative in every run ($-2.66$, $-1.20$, $-3.73$). Detector noise widens the intervals but does not explain the direction of the change.

\textbf{What the transform preserves.} Measured from the paired documents: 96.0\% of values replaced more than once in a document (459/478) receive the same surrogate each time; the character-class pattern (letters, digits, punctuation) is kept for 86.7\% of identifiers and phone numbers and 75.1\% of dates; 74.5\% of e-mail addresses are format-preserving x-masked (\tp{nxxxxx@\dots}) rather than replaced. An LLM judge rates 73.6\% of replaced names and places as fitting the document's language and region, against 85.4\% for the original values; this judge is not validated (Appendix~\ref{app:transform}). The vendor does not document its date-shift or locale policy, and we do not infer one.

\section{Discussion}
\textbf{What the evidence supports.} On the spans this transform replaces, substitution changes detectability by about one point, which is within $\pm2$ points pooled under overlap matching, but not for every benchmark, detector or matching criterion. The residual is not a single defect: well-formed surrogates are also missed, mostly because detection depends on the particular value drawn. Diagnosis classification was unaffected by substitution, Faker surrogates or redaction.

\textbf{Practical recommendations.} These follow from detectability evidence only. (1)~Prefer same-type, well-formed, locale-appropriate surrogates sampled from realistic distributions; maximally canonical values (as Faker emits) are easier to detect than real PHI, which works against hiding in plain sight. (2)~Format-preserving masking (\tp{nxxxxx@\dots}) is reasonable for direct identifiers such as IDs and e-mail addresses, where a realistic surrogate could collide with a real person; it trades some detectability for a stronger guarantee. (3)~Keep values that carry clinical meaning, such as age, consistent with the record; random ages broke age-group consistency in our downstream test. (4)~Redaction remains appropriate when downstream use does not need the entity signal. (5)~Validate on one's own data and detector: pooled equivalence did not hold for identifier-dense non-English text or for several detectors.

\textbf{The protocol.} The paired masked-span design, clustering-aware equivalence bounds and mechanism analyses need only black-box access to a transform and can audit any structure-preserving de-identifier.

\section*{Limitations}
Detectability preservation is a proxy: it does not measure distributional fidelity of surrogates, temporal consistency of shifted dates, or most downstream tasks; our downstream check covers one task family with small local models. Coverage is incomplete on every benchmark, and conclusions about substitution apply only to the spans the transform replaces, which are easier to detect than the spans it leaves. Gold spans are re-projected heuristically because the transform's replacement offsets were unavailable. The $\pm2$-point margin is a convention. C1 and C2 cover four of the eleven detectors and 120 documents per benchmark. LLM detectors are prompt-sensitive~\citep{gendron2024abstract,bao2024robustness}; we use one fixed prompt per model. Stronger reasoning strategies (\S2) may change detector behaviour and are outside our scope. All benchmarks are public or synthetic, and validation on real EHR text is future work.

\section*{Ethics and Conflict of Interest}
No real patient data is used. The authors are affiliated with Custodian Labs, which develops the transform evaluated here, and the work was conducted with Custodian Labs' support. To limit bias we frame the contribution as a protocol, include open baselines that reproduce without proprietary access, report results that qualify our claims (non-uniform equivalence, incomplete coverage), and release all predictions and scripts.

\bibliography{refs}

\clearpage
\appendix

\section{Detector Panel}\label{app:detectors}
\begin{table}[ht]\centering\small
\caption{Detector panel. All LLMs use the same zero-shot JSON span prompt and string-search span recovery.}\label{tab:panel}
\begin{adjustbox}{max width=\columnwidth}
\begin{tabular}{lll}
\toprule
Family & Detector & Notes\\
\midrule
Rule-based & Presidio (spaCy \tp{lg} models) & per-language NER + recognisers\\
Encoder tagger & OBI \tp{deid\_roberta\_i2b2} & RoBERTa, i2b2 2014 labels\\
LLM, local & Gemma 4 31B, Gemma 4 E4B & bf16, greedy\\
 & Qwen 3.5 4B, 9B, 35B-A3B & bf16, greedy, thinking off\\
 & Llama 3.1 8B, Llama 3.3 70B & bf16, greedy\\
 & DeepSeek V2-Lite & bf16, greedy\\
LLM, API & GPT-4o-mini & OpenAI API, temperature 0\\
\midrule
Subset only & GPT-5 & OpenAI API, default temperature\\
\bottomrule
\end{tabular}
\end{adjustbox}
\end{table}

\section{Margins}\label{app:margins}
\begin{figure}[ht]
\centering
\includegraphics[width=\linewidth]{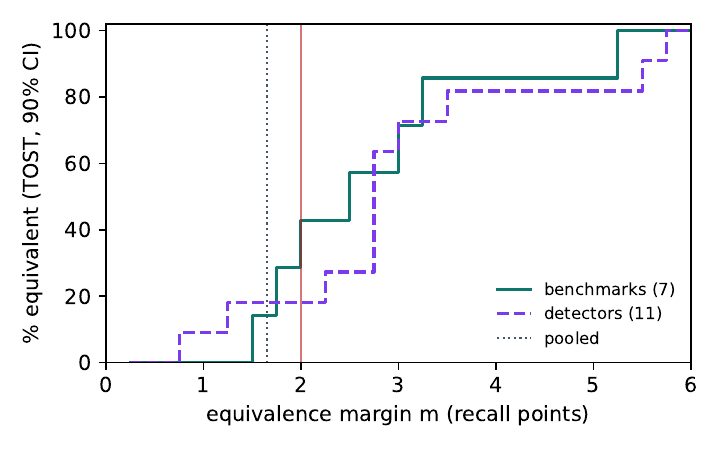}
\caption{Share of the seven benchmarks and eleven detectors whose 90\% cluster-bootstrap CI lies inside $(-m,m)$, as a function of the margin $m$. The dotted line marks the pooled $m^\ast=1.65$.}
\end{figure}

\section{Span Selection}\label{app:selection}
\begin{table}[ht]\centering\small
\caption{Recall on the original documents for gold spans the transform replaced vs.\ left in place, by coarse type (span$\times$detector outcomes).}\label{tab:selection}
\begin{adjustbox}{max width=\columnwidth}
\begin{tabular}{lrrrr}
\toprule
Gold type & Masked outc. & Recall & Unmasked outc. & Recall\\
\midrule
Name & 10{,}836 & 79.1 & 14{,}204 & 72.3\\
Date & 13{,}818 & 83.5 & 6{,}167 & 43.0\\
Location & 13{,}107 & 68.0 & 24{,}674 & 60.3\\
Identifier & 6{,}844 & 83.3 & 16{,}855 & 79.3\\
Contact & 8{,}930 & 77.5 & 7{,}896 & 71.4\\
Age & 2{,}265 & 60.7 & 3{,}390 & 60.6\\
Sex & 500 & 11.0 & 7{,}735 & 8.1\\
Other & 735 & 33.6 & 4{,}823 & 27.5\\
\midrule
All & & 75.9 & & 59.3\\
\bottomrule
\end{tabular}

\end{adjustbox}
\end{table}

\section{GPT-5 on a Stratified Subset}\label{app:gpt5}
We ran GPT-5~\citep{openai2025gpt5} (\tp{gpt-5-2025-08-07}) with the same prompt and parser on 199 documents (up to 30 per benchmark, all seven benchmarks), at its default temperature because it does not accept temperature 0. On these documents GPT-5 has the highest whole-document F1 of the panel (Table~\ref{tab:gpt5}). The masked-span change is computed on the 173 documents that GPT-5, GPT-4o-mini and Gemma~4~31B all cover (678 replaced spans), so its intervals are wide. The panel's OpenAI detector is GPT-4o-mini: re-running 20 documents with the panel's code reproduced the stored span sets exactly for 17/20 with GPT-4o-mini and 2/20 with GPT-5.

\begin{table}[ht]\centering\small
\caption{GPT-5 and selected panel detectors on the same 199 documents. $^\dagger$182 documents.}\label{tab:gpt5}
\begin{adjustbox}{max width=\columnwidth}
\begin{tabular}{lrrr}
\toprule
Detector & F1 (exact) & F1 (overlap) & Masked $\Delta$ [90\% CI]\\
\midrule
GPT-5 & .787 & .853 & $-$0.44 [$-$2.27, $+$1.41]\\
Llama 3.3 70B$^\dagger$ & .774 & .842 & --\\
Gemma 4 31B & .765 & .841 & $+$0.74 [$-$1.15, $+$2.83]\\
GPT-4o-mini & .749 & .824 & $-$3.98 [$-$6.22, $-$1.79]\\
Qwen 3.5 35B-A3B & .729 & .820 & --\\
\bottomrule
\end{tabular}

\end{adjustbox}
\end{table}

\section{Annotation Protocol}\label{app:annotation}
Raters saw the original value, the surrogate, the gold type, and a 70-character context window around each in $d$ and $d'$, and chose one category: \emph{well-formed} (plausible, same type, fits language and context); \emph{malformed or truncated} (fragment, non-word or impossible value); \emph{x-masked}; \emph{type-inconsistent} (well formed but a different kind of entity, e.g.\ a country becoming a nationality); \emph{locale-inconsistent}; \emph{context-inconsistent} (contradicts the clinical or document context, e.g.\ the patient's sex or stated age); \emph{salience change} (a well-known entity replaced by an obscure one); or \emph{other/uncertain}. Surrogates were re-extracted with the corrected re-projection. The earlier AI labels had counted 15 alignment artefacts as truncations.

\section{Transform Properties}\label{app:transform}
x-masking rates by benchmark: ASQ-PHI 5.2\%, MEDDOCAN 8.0\%, PII-en 9.8\%, PII-de 11.3\%, PII-nl 15.9\%, PII-fr 16.5\% of replaced spans. The locale judge (Qwen~3.5~9B, option likelihood) answers whether a value is a name or place typical for a document written in the benchmark's language; it rates 73.6\% of commercial surrogates, 85.4\% of the original values and 87.8\% of locale-matched Faker values as consistent. The judge also rates many genuine names in PII-Masking-300k as atypical, so these numbers are descriptive only; human raters found locale inconsistency in under 1\% of the annotated surrogates.

\section{Worked Examples}\label{app:examples}
\begin{table*}[ht]\centering\small
\caption{Worked transform examples. Only the sensitive value moves; abbreviations, non-English syntax and JSON structure are preserved. In the JSON case the username is left unchanged, a coverage decision.}\label{tab:examples}
\begin{tabular}{p{2.3cm}p{6.0cm}p{6.0cm}}
\toprule
Case & Original & Transformed\\
\midrule
ASQ-PHI (query) & \dots a 34-year-old female \dots like \oldv{Anna~S.}, \dots at Methodist Hospital on \oldv{April~12,~2023}? & \dots a 35-year-old female \dots like \newv{Maria~S.}, \dots at Methodist Hospital on \newv{March~13,~2021}?\\
\addlinespace
ASQ-PHI (shorthand) & Rec mgmt of 70yo M w/ CHF, seen by Dr.~\oldv{John~L.} at \oldv{Mt.~Sinai} on \oldv{Feb~21,~2023}. alt tx options\dots & Rec mgmt of 73yo M w/ CHF, seen by Dr.~\newv{James~L.} at \newv{Mt.~Egypt} on \newv{Nov~19,~2021}. alt tx options\dots\\
\addlinespace
PII (German) & \dots erm\"achtigen hiermit \oldv{Monsignore}, \dots Mit Datum \oldv{23/07/2011}\dots & \dots erm\"achtigen hiermit \newv{Fulgenzio}, \dots Mit Datum \newv{24/07/2011}\dots\\
\addlinespace
PII (French, JSON) & \tp{\char`\{"Date":"\oldv{20/05/2022}", "City":"Saint-Priest", "Username":"phprosdocimo"\char`\}} & \tp{\char`\{"Date":"\newv{21/05/2023}", "City":"Saint-Priest", "Username":"phprosdocimo"\char`\}}\\
\bottomrule
\end{tabular}
\end{table*}

\textbf{All detectors on one transformed document.} Predictions of the eleven detectors on \tp{asq\_00001}, sorted by this document's F1 (\hg{green}: caught, \hy{yellow}: false positive, \hr{red}: missed). The values shown are surrogates produced by the transform. These panels are illustrative and use the uncorrected alignment.

{\footnotesize
\setlength{\fboxsep}{1pt}%
\newcommand{\panel}[3]{\par\vspace{2.5pt}\noindent\fcolorbox{cline}{white}{\parbox{0.965\linewidth}{%
\textbf{#1}\hfill{\scriptsize\color{cmuted}#2}\\[2pt]\ttfamily\scriptsize #3}}}
\panel{Gold (reference)}{4 PHI spans}{Evaluation of long-term outcomes for bypass surgery in patients over 60, referencing Mr. \hg{John T.}, operated at \hg{Saint. Vxxxxxxxx} on \hg{April 25th, 2018} (ID: \hg{987654321})?}
\panel{GPT-4o-mini}{tp 4 $\cdot$ fp 0 $\cdot$ missed 0 $\cdot$ F1 100\%}{Evaluation of long-term outcomes for bypass surgery in patients over 60, referencing \hg{Mr. John T.}, operated at \hg{Saint. Vxxxxxxxx} on \hg{April 25th, 2018} (\hg{ID: 987654321})?}
\panel{Llama 3.1-8B}{tp 4 $\cdot$ fp 0 $\cdot$ missed 0 $\cdot$ F1 100\%}{Evaluation of long-term outcomes for bypass surgery in patients over 60, referencing Mr. \hg{John T.}, operated at \hg{Saint. Vxxxxxxxx} on \hg{April 25th, 2018} (ID: \hg{987654321})?}
\panel{Qwen 3.5-4B}{tp 4 $\cdot$ fp 0 $\cdot$ missed 0 $\cdot$ F1 100\%}{Evaluation of long-term outcomes for bypass surgery in patients over 60, referencing \hg{Mr. John T.}, operated at \hg{Saint. Vxxxxxxxx} on \hg{April 25th, 2018} (ID: \hg{987654321})?}
\panel{Gemma 4 31B}{tp 4 $\cdot$ fp 0 $\cdot$ missed 0 $\cdot$ F1 100\%}{Evaluation of long-term outcomes for bypass surgery in patients over 60, referencing \hg{Mr. John T.}, operated at \hg{Saint. Vxxxxxxxx} on \hg{April 25th, 2018} (ID: \hg{987654321})?}
\panel{Llama 3.3-70B}{tp 4 $\cdot$ fp 0 $\cdot$ missed 0 $\cdot$ F1 100\%}{Evaluation of long-term outcomes for bypass surgery in patients over 60, referencing Mr. \hg{John T.}, operated at \hg{Saint. Vxxxxxxxx} on \hg{April 25th, 2018} (ID: \hg{987654321})?}
\panel{Qwen 3.5-35B-A3B}{tp 4 $\cdot$ fp 0 $\cdot$ missed 0 $\cdot$ F1 100\%}{Evaluation of long-term outcomes for bypass surgery in patients over 60, referencing \hg{Mr. John T.}, operated at \hg{Saint.} \hg{Vxxxxxxx}x on \hg{April 25th}, \hg{2018} (ID: \hg{987654321})?}
\panel{Gemma 4 E4B}{tp 4 $\cdot$ fp 0 $\cdot$ missed 0 $\cdot$ F1 100\%}{Evaluation of long-term outcomes for bypass surgery in patients over 60, referencing Mr. \hg{John T.}, operated at \hg{Saint. Vxxxxxxxx} on \hg{April 25th, 2018} (ID: \hg{987654321})?}
\panel{Qwen 3.5-9B}{tp 4 $\cdot$ fp 0 $\cdot$ missed 0 $\cdot$ F1 100\%}{Evaluation of long-term outcomes for bypass surgery in patients over 60, referencing \hg{Mr. John T.}, operated at \hg{Saint. Vxxxxxxxx} on \hg{April 25th, 2018} (ID: \hg{987654321})?}
\panel{OBI deid\_roberta}{tp 4 $\cdot$ fp 1 $\cdot$ missed 0 $\cdot$ F1 89\%}{Evaluation of long-term outcomes for bypass surgery in patients over \hy{60}, referencing Mr. \hg{John} \hg{T}., operated at \hg{Saint}. \hg{Vxxxxxxxx} on \hg{April 25th,} \hg{2018} (ID: \hg{987654321})?}
\panel{Microsoft Presidio}{tp 3 $\cdot$ fp 1 $\cdot$ missed 1 $\cdot$ F1 75\%}{Evaluation of long-term outcomes for bypass surgery in patients over \hy{60}, referencing Mr. \hg{John T.}, operated at \hr{Saint. Vxxxxxxxx} on \hg{April 25th, 2018} (ID: \hg{987654321})?}
\panel{DeepSeek V2-Lite}{tp 2 $\cdot$ fp 0 $\cdot$ missed 2 $\cdot$ F1 67\%}{Evaluation of long-term outcomes for bypass surgery in patients over 60, referencing Mr. \hg{John T.}, operated at \hr{Saint. Vxxxxxxxx} on \hg{April 25th, 2018} (ID: \hr{987654321})?}

\par}

\textbf{A dense Spanish clinical header (MEDDOCAN).} On identifier-dense non-English text the detectors diverge far more; the garbled surrogate \tp{Cuidad Real} (from \tp{Ciudad Real}) is still caught by most of them.

{\footnotesize
\setlength{\fboxsep}{1pt}%
\newcommand{\panel}[3]{\par\vspace{2.5pt}\noindent\fcolorbox{cline}{white}{\parbox{0.965\linewidth}{%
\textbf{#1}\hfill{\scriptsize\color{cmuted}#2}\\[2pt]\ttfamily\scriptsize #3}}}
\panel{Gold (reference)}{12 PHI spans}{\hg{Cuidad Real}. CP: \hg{13002}. Datos asistenciales. Referencia de nacimiento: \hg{24/10/1963}. País: \hg{España}. Mejores: \hg{53 años} Sexo: \hg{H}. Referencia de Ingreso: \hg{14/12/2017}. Episodio: \hg{746589123}. Médico: \hg{Luis Ruiz Camuñas} N\textordmasculine Col: \hg{03 14 16485}. Historia Actual: \hg{Varón} de \hg{53 años} q}
\panel{DeepSeek V2-Lite}{tp 11 $\cdot$ fp 0 $\cdot$ missed 1 $\cdot$ F1 96\%}{\hg{Cuidad Real}. CP: \hg{13002}. Datos asistenciales. Referencia de nacimiento: \hg{24/10/1963}. País: \hg{España}. Mejores: \hg{53 años} Sexo: \hg{H}. Referencia de Ingreso: \hg{14/12/2017}. Episodio: \hg{746589123}. Médico: \hg{Luis Ruiz Camuñas} N\textordmasculine Col: \hr{03 14 16485}. Historia Actual: \hg{Varón de 53 años q}}
\panel{Qwen 3.5-4B}{tp 11 $\cdot$ fp 0 $\cdot$ missed 1 $\cdot$ F1 96\%}{\hg{Cuidad Real.} CP: \hg{13002.} Datos asistenciales. Referencia de nacimiento: \hg{24/10/1963.} País: \hg{España.} Mejores: \hg{53} años Sexo: \hg{H.} Referencia de Ingreso: \hg{14/12/2017.} Episodio: \hg{746589123.} Médico: \hg{Luis Ruiz Camuñas} N\textordmasculine Col: \hg{03} 14 \hg{16485.} Historia Actual: \hr{Varón} de \hg{53} años q}
\panel{OBI deid\_roberta}{tp 10 $\cdot$ fp 0 $\cdot$ missed 2 $\cdot$ F1 91\%}{\hg{Cu}idad \hg{Real}. CP: \hg{13}002. Datos asistenciales. Referencia de nacimiento: \hg{24/10/1963}. País: \hg{Esp}aña. Mejores: \hg{53} años Sexo: \hr{H}. Referencia de Ingreso: \hg{14/12/2017}. Episodio: \hg{746589123}. Médico: \hg{Luis Ruiz} \hg{Camuñ}as N\textordmasculine Col: \hg{03 14} \hg{16485}. Historia Actual: \hr{Varón} de \hg{53} años q}
\panel{Qwen 3.5-9B}{tp 10 $\cdot$ fp 0 $\cdot$ missed 2 $\cdot$ F1 91\%}{\hg{Cuidad Real}. CP: \hg{13002}. Datos asistenciales. Referencia de nacimiento: \hg{24/10/1963}. País: \hg{España}. Mejores: \hg{53} años Sexo: \hr{H}. Referencia de Ingreso: \hg{14/12/2017}. Episodio: \hg{746589123}. Médico: \hg{Luis Ruiz Camuñas} N\textordmasculine Col: \hg{03 14 16485}. Historia Actual: \hr{Varón} de \hg{53} años q}
\panel{Gemma 4 31B}{tp 9 $\cdot$ fp 0 $\cdot$ missed 3 $\cdot$ F1 86\%}{\hg{Cuidad Real}. CP: \hg{13002}. Datos asistenciales. Referencia de nacimiento: \hg{24/10/1963}. País: \hr{España}. Mejores: \hg{53 años} Sexo: \hr{H}. Referencia de Ingreso: \hg{14/12/2017}. Episodio: \hg{746589123}. Médico: \hg{Luis Ruiz Camuñas} N\textordmasculine Col: \hg{03 14 16485}. Historia Actual: \hr{Varón} de \hg{53 años} q}
\panel{Qwen 3.5-35B-A3B}{tp 9 $\cdot$ fp 1 $\cdot$ missed 3 $\cdot$ F1 82\%}{\hg{Cuidad Real}. CP: \hg{13002}. Datos asistenciales. Referencia de nacimiento: \hg{24/10/1963}. País: \hg{España}. Mejores: \hg{53} años Sexo: \hr{H}. Referencia de Ingreso: \hg{14/12/2017}. Episodio: \hg{746589123}. Médico: \hg{Luis Ruiz Camuñas} \hy{N\textordmasculine }Col: \hg{03 14 16485}. Historia Actual: \hr{Varón} de \hr{53 años} q}
\panel{GPT-4o-mini}{tp 8 $\cdot$ fp 0 $\cdot$ missed 4 $\cdot$ F1 80\%}{\hg{Cuidad Real}. CP: \hg{13002}. Datos asistenciales. Referencia de nacimiento: \hg{24/10/1963}. País: \hr{España}. Mejores: \hg{53} años Sexo: \hr{H}. Referencia de Ingreso: \hg{14/12/2017}. Episodio: \hg{746589123}. Médico: \hg{Luis Ruiz Camuñas} N\textordmasculine Col: \hg{03 14 16485}. Historia Actual: \hr{Varón} de \hr{53 años} q}
\panel{Llama 3.3-70B}{tp 8 $\cdot$ fp 0 $\cdot$ missed 4 $\cdot$ F1 80\%}{\hg{Cuidad Real}. CP: \hg{13002}. Datos asistenciales. Referencia de nacimiento: \hg{24/10/1963}. País: \hr{España}. Mejores: \hg{53} años Sexo: \hr{H}. Referencia de Ingreso: \hg{14/12/2017}. Episodio: \hg{746589123}. Médico: \hg{Luis Ruiz Camuñas} N\textordmasculine Col: \hg{03 14 16485}. Historia Actual: \hr{Varón} de \hr{53 años} q}
\panel{Gemma 4 E4B}{tp 8 $\cdot$ fp 0 $\cdot$ missed 4 $\cdot$ F1 80\%}{\hg{Cuidad Real}. CP: \hg{13002}. Datos asistenciales. Referencia de nacimiento: \hg{24/10/1963}. País: \hr{España}. Mejores: \hg{53} años Sexo: \hr{H}. Referencia de Ingreso: \hg{14/12/2017}. Episodio: \hr{746589123}. Médico: \hg{Luis Ruiz Camuñas} N\textordmasculine Col: 03 14 1\hg{6}485. Historia Actual: \hr{Varón} de \hg{53} años q}
\panel{Microsoft Presidio}{tp 6 $\cdot$ fp 1 $\cdot$ missed 6 $\cdot$ F1 63\%}{\hg{Cuidad Real}. CP: \hr{13002}. Datos asistenciales. Referencia de nacimiento: \hg{24/10/1963}. \hy{País}: \hg{España}. Mejores: \hr{53 años} Sexo: \hr{H}. Referencia de Ingreso: \hg{14/12/2017}. Episodio: \hr{746589123}. Médico: \hg{Luis Ruiz Camuñas N\textordmasculine Col}: \hg{03 14 16485}. Historia Actual: \hr{Varón} de \hr{53 años} q}
\panel{Llama 3.1-8B}{tp 0 $\cdot$ fp 0 $\cdot$ missed 12 $\cdot$ F1 0\%}{\hr{Cuidad Real}. CP: \hr{13002}. Datos asistenciales. Referencia de nacimiento: \hr{24/10/1963}. País: \hr{España}. Mejores: \hr{53 años} Sexo: \hr{H}. Referencia de Ingreso: \hr{14/12/2017}. Episodio: \hr{746589123}. Médico: \hr{Luis Ruiz Camuñas} N\textordmasculine Col: \hr{03 14 16485}. Historia Actual: \hr{Varón} de \hr{53 años} q}

\par}

\end{document}